\documentclass{ieeeaccess}
\usepackage{cite}
\usepackage{amsmath,amssymb,amsfonts}
\usepackage{algorithmic}
\usepackage{graphicx}
\usepackage{textcomp}

\usepackage{multirow}

\def\BibTeX{{\rm B\kern-.05em{\sc i\kern-.025em b}\kern-.08em
    T\kern-.1667em\lower.7ex\hbox{E}\kern-.125emX}}
\begin{document}
\history{Date of publication xxxx 00, 0000, date of current version xxxx 00, 0000.}
\doi{10.1109/ACCESS.2017.DOI}

\title{A Random Ensemble of Encrypted Vision Transformers for Adversarially Robust Defense
}
\author{\uppercase{Ryota Iijima},
\uppercase{Sayaka Shiota, \IEEEmembership{Member, IEEE}},
\uppercase{Hitoshi Kiya, \IEEEmembership{Life Fellow, IEEE}}}
\address{Department of Computer Science, Tokyo Metropolitan University, Tokyo, 191-0065, Japan}

\markboth
{Author \headeretal: Preparation of Papers for IEEE TRANSACTIONS and JOURNALS}
{Author \headeretal: Preparation of Papers for IEEE TRANSACTIONS and JOURNALS}

\corresp{Corresponding author: Hitoshi Kiya (e-mail: kiya@ed.tmu.ac.jp).}

\begin{abstract}
Deep neural networks (DNNs) are well known to be vulnerable to adversarial examples (AEs).
In previous studies, the use of models encrypted with a secret key was demonstrated to be robust against white-box attacks, but not against black-box ones.
In this paper, we propose a novel method using the vision transformer (ViT) that is a random ensemble of encrypted models for enhancing robustness against both white-box and black-box attacks.
In addition, a benchmark attack method, called AutoAttack, is applied to models to test adversarial robustness objectively.
In experiments, the method was demonstrated to be robust against not only white-box attacks but also black-box ones in an image classification task on the CIFAR-10 and ImageNet datasets.
The method was also compared with the state-of-the-art in a standardized benchmark for adversarial robustness, RobustBench, and it was verified to outperform conventional defenses in terms of clean accuracy and robust accuracy.
\end{abstract}


\titlepgskip=-15pt

\maketitle

\section{Introduction}
Deep neural networks (DNNs) have achieved great success in various computer vision tasks and have been deployed in many applications including security-critical ones, such as face recognition and object detection for autonomous driving.
However, DNNs are known to be vulnerable to adversarial examples (AEs), which fool DNNs by adding small perturbations to input images without any effect on human perception.
In addition, AEs designed for a source model can deceive other (target) models. This property, called transferability, makes it easy to mislead various DNN models.
This is an urgent issue that has a negative impact on the reliability of applications using DNNs. \par

In previous studies, various methods were proposed to build models robust against AEs.
Adversarial training \cite{goodfellow2015explaining,kurakin2017adversarial, Hongyang2019theoratically,carmon2019unlabeled} is widely known as a means of defense against AEs, where AEs are used as a part of training data to improve robustness against AEs.
However, it has the problem of degraded model performance that occurs when test data is clean. \par

Another approach for constructing robust models is to train models by using images encrypted with secret keys \cite{maung2020encryption, aprilpyone2021block}.
The encrypted models have been verified to be robust against white-box attacks if the keys are not known to adversaries.
Furthermore, the approach is effective in avoiding the influence of transferability \cite{tanaka2022on}.
However, this approach is still vulnerable to black-box attacks because these attacks do not need to know any secret key. \par

Accordingly, we propose a random ensemble of encrypted vision transformers (ViTs) that is inspired by perceptual image encryption \cite{chuman2019encryption, watanabe2004a}, model encryption \cite{maung2020encryption,aprilpyone2021block,ito2021image,tanaka2022on,april2022priavcy} and an ensemble of models \cite{pang2019improving, yang2020dverge}.
An ensemble of encrypted models was discussed as one type of adversarial defense \cite{aprilpyone2021block,maungmaung2021ensemble}, but the ensemble model is not yet robust against black-box attacks when it is used as the source model.
The proposed method with a random ensemble allows us not only to use models robust against black-box attacks but
to also utilize ones robust against white-box attacks.
In  experiments, the effectiveness of the proposed method is verified on the CIFAR-10 and ImageNet datasets by using a benchmark attack, AutoAttack \cite{croce2020reliable} \par

We reported that a random ensemble of encrypted sub-models was effective for enhancing robustness against AEs in a preliminary experiment as the first report \cite{iijima2023enhanced}.
The report is extended this time by adding additional experiments and detailed considerations including comparison with the state-of-the-art and discussion on the ImageNet dataset.
The rest of this paper is structured as follows.
Section \ref{sec:related} presents related work on adversarial examples, defense methods, and the vision transformer.
Regarding the proposed method, Section \ref{sec:method} includes an overview, threat model, and a random ensemble of encrypted models.
Experiments for verifying the effectiveness of the method, including classification accuracy, effects of the number of sub-models, and comparison with the state-of-the-art, are presented in Section \ref{sec:experiment}, and Section \ref{sec:conlusion} concludes this paper.

\section{Related work} \label{sec:related}
\subsection{Adversarial Examples}\label{sec:ae}

AEs are used to mislead machine learning models \cite{christian2014intriguing, biggio2013evasion, goodfellow2015explaining}.
Traditionally, adversarial attacks are classified into three types in accordance with the knowledge of a particular model and training data available to the adversary: white-box, black-box, and gray-box.
Under white-box settings \cite{goodfellow2015explaining, madry2018towards, croce2020minimally}, the adversary has direct access to the model, its parameters, training data, and defense mechanism.
In contrast, the adversary does not have any knowledge on the model, except the output of the model in black-box attacks \cite{su2019one, andriushchenko2020square, li2019nattack}.
Between white-box and black-box attacks, there are gray-box ones that imply that the adversary knows something about the system \cite{vivek2018gray, xiang2021side}. \par

The model used for the design of AEs is called a source model, and the model that is the final objective of the attack is called a target model. 
The source model is the same as the target model in general.
However, even if the source model is different from the target model, AEs may deceive the target model \cite{papernot2016transferability, christian2014intriguing, liu2017delving, mahmood2021robustness}.
This property is called adversarial transferability.
Therefore, several black-box and gray-box attacks prepare substitute models to generate AEs. \par
AEs are also classified into two types depending on how they are created: perturbation-based AEs and unrestricted AEs.
Perturbations in perturbation-based AEs are restricted by matrix p-norms such as ${\ell}_{\infty}$ \cite{madry2018towards}, ${\ell}_{2}$ \cite{moosavi2016deepfool}, ${\ell}_{1}$ \cite{chen2018ead}, and ${\ell}_{0}$ \cite{papernot2016limitations} to be imperceptible to humans.
In contrast, unrestricted AEs \cite{brown2018unrestricted} are crafted by using special transformation \cite{engstrom2019exploring} or generative models \cite{song2018advances}.
In this paper, we use perturbation-based (${\ell}_{\infty}$-norm bounded) AEs to evaluate the proposed method because they are used in a benchmark attack method called AutoAttack \cite{croce2020reliable}.

\subsection{Defense Method}\label{sec:defense}
Various adversarial defenses have been studied to construct models robust against AEs so far.
There are two strategies for adversarial defenses. \par
The first strategy aims to build models that directly classify AEs without removing perturbations.
Adversarial training \cite{goodfellow2015explaining,kurakin2017adversarial, Hongyang2019theoratically,carmon2019unlabeled}, which trains models with AEs, is one such widely known strategy.
Madry et al. approach adversarial training as an optimization problem, and they utilize the projected gradient descent (PGD) adversary as a universal attack one to craft AEs \cite{madry2018towards}. 
While adversarial training with the PGD adversary is high in robustness, the required computational resources are also high.
Many studies have made progress to reduce the computational cost, such as free-adversarial training \cite{shafahi2019adversarial}, fast adversarial training \cite{Wong2020Fast}, and single-step adversarial training \cite{de2022make}.
However, it was shown that models trained with AEs under the ${\ell}_{\infty}$-norm can still be vulnerable to ${\ell}_{1}$ norm-bounded AEs \cite{sharma2017attacking}.
In contrast, certified defenses are also one type of defense in which perturbation is not removed.
Certified defenses provide strict mathematical guarantees of robustness against AEs in such a way that there are no AEs within some bounds \cite{wong2018provable,raghunathan2018certified,cohen2019certified,hein2017formal}.
However, these defenses do not work well against some types of perturbation such as generative perturbation \cite{poursaeed2018generative} and parametric perturbation \cite{liu2018beyond} \par

The second strategy is to pre-process input images to reduce the effect of perturbations in AEs.
In previous studies, various transformation methods were proposed, such as thermometer encoding \cite{jacob2018thermometer}, diverse image processing techniques \cite{guo2018countering, xie2018mitigating}, denoising strategies \cite{liao2018defense, niu2020limitations}, and GAN-based transformation \cite{song2018pixeldefend}.
At first, the performance of these input transformation-based defenses was high, but it was seen that they were vulnerable to adaptive attacks \cite{athalye2018obfuscated, tramer2017ensemble}.
Key-based defenses are also one type of defense method belonging to the second strategy.
Unlike other defenses, they utilize secret keys to defend models so that they have an information advantage over adversarial attacks \cite{Taran2018bridging, aprilpyone2021block, hitoshi2022an, maung2020encryption}.
Unless secret keys are leaked, adversarial attacks do not break key-based defenses. \par

However, key-based attacks are not robust against black-box attacks such as Square Attack \cite{andriushchenko2020square}.
In this paper, we aim to solve the issue that conventional key-based defenses have.

\subsection{Properties of Vision Transformer}
Transformer-based models have been widely used in natural language processing (NLP) tasks \cite{devlin2019bert}.
Inspired by the success of NLP tasks, the vision transformer (ViT) was proposed for computer vision tasks \cite{dosovitskiy2021an}.
We focus on the following properties of ViT that enhance the robustness of models against AEs.
\begin{enumerate}
\renewcommand{\labelenumi}{(\alph{enumi})}
    \item ViT provides high performance in image classification tasks, compared with CNN models \cite{dosovitskiy2021an}.
    \item The transferability between ViT and CNN models is low \cite{mahmood2021robustness}.
    \item The transferability among ViT models encrypted with different keys is low \cite{tanaka2022on}.
    \item Isotropic networks including ViT have a high similarity with block-wise image encryption \cite{hitoshi2022an, qi2022privacy, nagamori2023domain}. 
\end{enumerate}
Properties (b) and (c) are important to robust models. 
If the transferability between models is high, effective AEs are easily designed even when the model parameters of a target model are not disclosed.
Property (d) means that it is easy to prepare encrypted models with high performance by using a block-wise encryption method.
The properties are obtained from the patch embedding structure of ViT.
Accordingly, it is expected that ViT models encrypted with a block-wise encryption method are robust against various attacks under some requirements.

\section{Methodology} \label{sec:method}

\subsection{Overview}
Fig. \ref{fig:framework} shows an overview of the proposed method.
The method considers satisfying the two requirements below.
\begin{itemize}
    \item It is robust against various AEs.
    \item It has a high classification accuracy even when inputting test images without any adversarial noise.
\end{itemize}
To fulfill the above requirements, we propose a random ensemble of ViTs in this paper. \par
As shown in the figure, each training image is encrypted to generate $N$ encrypted images by using $N$ keys $K=\{ K_{1},\dots,K_{N} \}$.
$N$ encrypted sub-models are trained by using encrypted images, and an ensemble of the sub-models is constructed.
For testing, $N$ test images encrypted with the keys used for training the sub-models are generated from a plain image, and they are input to the random
ensemble model to get an estimated result. 

\Figure[t][scale=0.25]{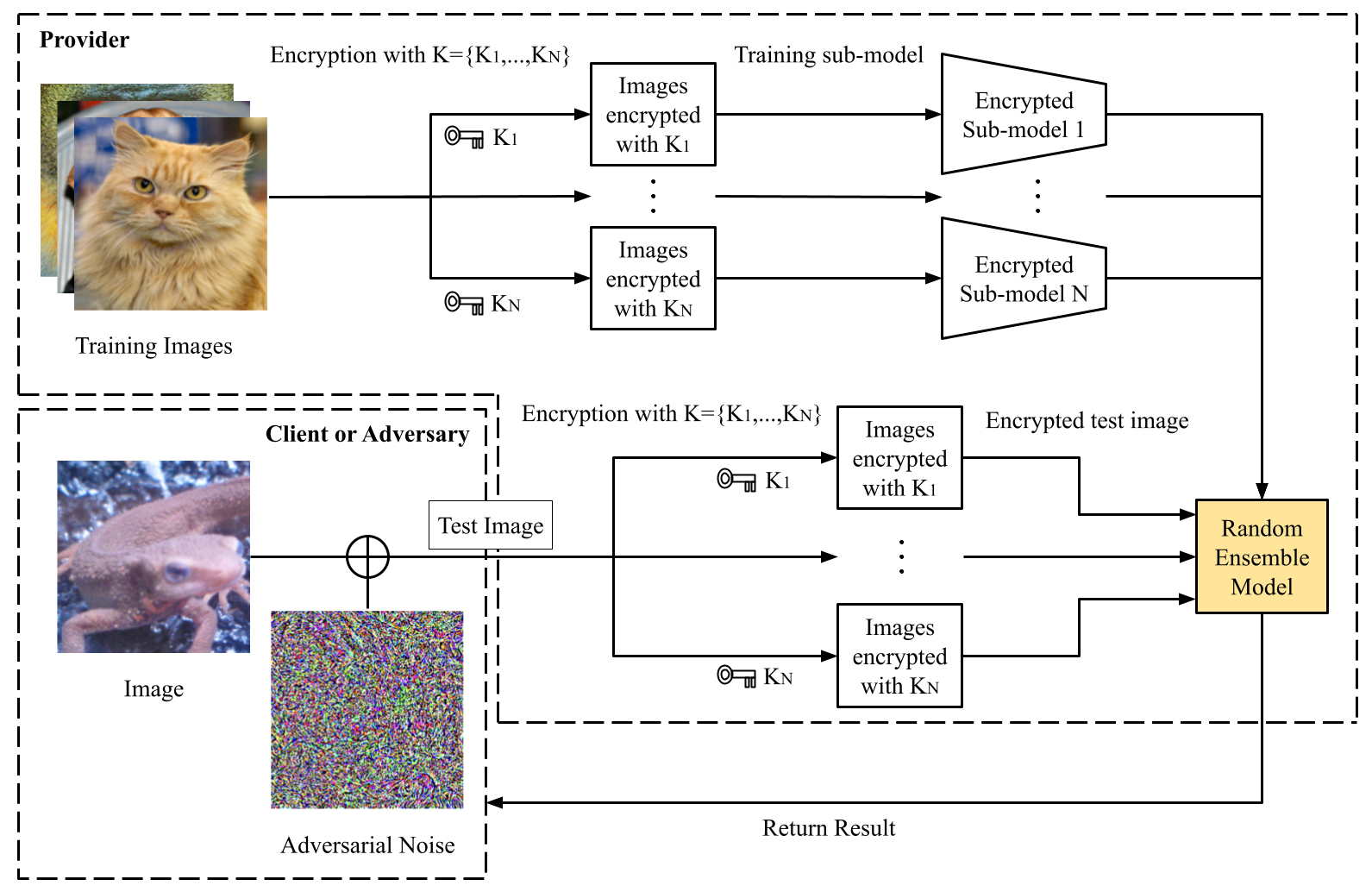}{Framework of proposed method. \label{fig:framework}}

\subsection{Threat Model}
The goal of an adversarial defense is to keep the classification accuracy on both clean images and adversarial examples high. To evaluate a defense method, precisely defining threat models is necessary.
A threat model includes a set of assumptions such as an adversary's goals and knowledge. \par
\noindent
\textbf{Adversary's Goals: }
An adversary can construct adversarial examples to achieve different goals when attacking a model: whether to reduce the performance accuracy (i.e., untargeted attacks) or to classify a targeted class (i.e., targeted attacks).
Formally, untargeted attacks will mislead a classifier to a wrong label, given an adversarial example, and targeted ones will force the classifier to a targeted label. \par
\noindent
\textbf{Adversary's Knowledge: }
The adversary's knowledge can be white-box (inner workings of the defense mechanism, complete knowledge on the model and its parameters), black-box (no knowledge on the model) and gray-box, that is, anything in between white-box and black-box. \par
In this paper, we focus on key-based defense.
We consider both white-box and black-box attacks while keeping a secret key. \par
\noindent
\textbf{Attack Scenarios: } Many defenses against AEs have been proposed, but it is very difficult to objectively evaluate defense methods without an independent test.
For this reason, AutoAttack \cite{croce2020reliable}, which is an ensemble of the adversarial attacks used to test adversarial robustness objectively, was proposed as a benchmark attack.
AutoAttack consists of four attack methods: Auto-PGD-cross entropy (APGD-ce) \cite{croce2020reliable}, APGD-target (APGD-t), FAB-target (FAB-t) \cite{croce2020minimally}, and Square Attack \cite{andriushchenko2020square}, which have the properties summarized in Table \ref{tab:autoattack}.
In this paper, we use these four attack methods and AutoAttack to objectively compare the proposed method with the state-of-the-art.

\renewcommand{\arraystretch}{1.1}
\begin{table}[t]
    \centering
    \caption{Attack methods used in AutoAttack}
    \small
    \begin{tabular}{c|c|c}
    \multirow{2}{*}{Attack} & White-box(W)/ & Target(T)/    \\
                            & Black-box(B)  & Non-target(N) \\
    \hline
    APGD-ce                 & W             & N             \\
    APGD-t                  & W             & T             \\
    FAB-t                   & W             & T             \\
    Square                  & B             & N             \\
    \end{tabular}
    \label{tab:autoattack}
\end{table}
\renewcommand{\arraystretch}{1}

\subsection{Notations}
The following notations are utilized throughout this paper.
\begin{itemize}
    \item $W$, $H$, and $C$ denote the width, height, and number of channels of an image, respectively.
    \item The tensor $x \in [0, 1]^{C \times W \times H}$ represents an input color image.
    \item The tensor $x' \in [0, 1]^{C \times W \times H}$ represents an encrypted image.
    \item $M$ is the block size of an image.
    \item $W_\mathrm{b} = \frac{W}{M}$ and $H_\mathrm{b} = \frac{H}{M}$ are the number of blocks across width $W$ and height $H$.
    We assume that $W$ and $H$ are divisible by $M$, so $W_\mathrm{b}$ and $H_\mathrm{b}$ are positive integers.
    \item $p_\mathrm{b} = C \times M \times M$ is the number of pixels in a block.
    \item $B_{w \, h} \in \mathbb{R}^{C \times M \times M}$ is a block in an image, where $w \in \{ 1, ..., W_\mathrm{b} \}$, $h \in \{ 1, ..., H_\mathrm{b} \}$.
    \item $b_{w \, h} \in \mathbb{R}^{p_\mathrm{b}}$ is a flattened version of block $B_{w \, h}$.
    \item A pixel value in $b_{w \, h}$ is denoted by $b_{w \, h}(c)$, where $c \in \{1, \dots, p_{\mathrm{b}}\}$.
    \item $K=\{ K_{1},\dots,K_{i},\dots,K_{N} \}$ is a set of secret keys, where $K_{i}$ is the key for the $i^{\mathrm{th}}$ sub-model, and $N$ is the number of sub-models.
\end{itemize}

\subsection{Random Ensemble}
A novel framework for adversarial defenses is proposed here.
Fig. \ref{fig:architecture} shows a random ensemble of encrypted sub-models in which each encrypted test image is input to the corresponding sub-model, and $S$ outputs are randomly selected from $N$ outputs, where $3 \leq S \leq N$.
An estimated result is finally decided by using $S$ outputs. \par

Image encryption, the training of sub-models with encrypted images, and the ensemble of sub-models and testing used in the framework are explained under the use of ViT with a patch size of $M \times M$ for sub-models below.

\Figure[t][scale=0.29]{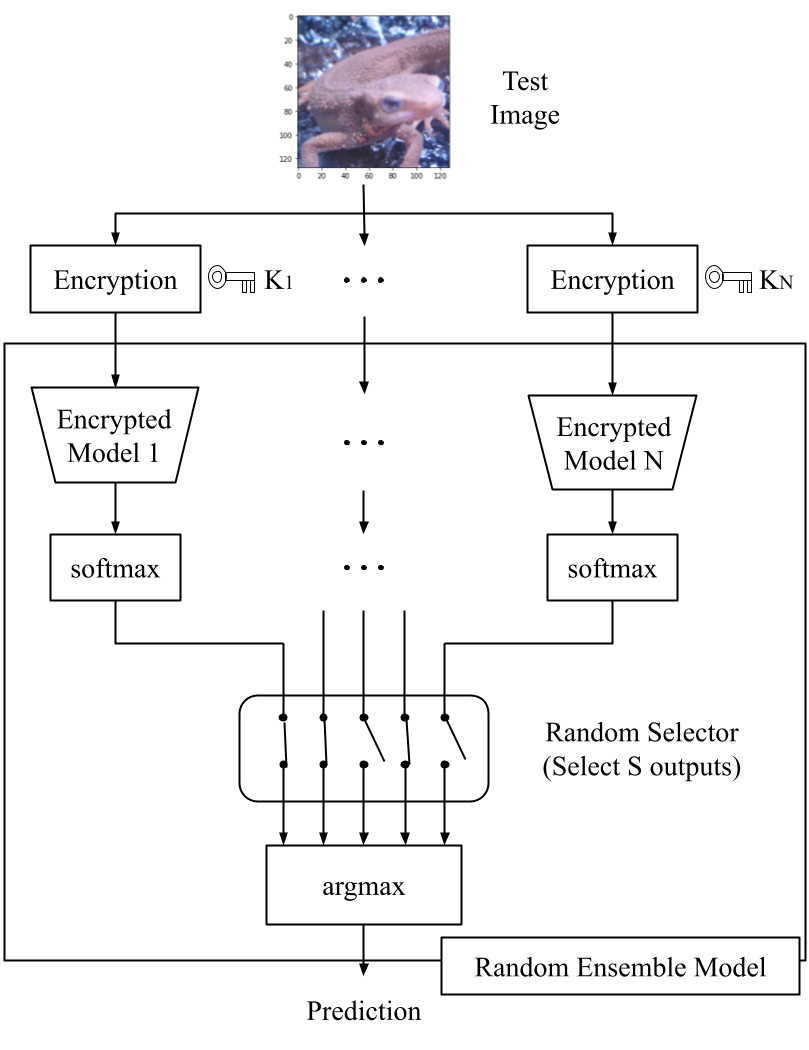}{Random ensemble of encrypted models. \label{fig:architecture}}

\par
\noindent
\textbf{Image Encryption} \par
Sub-models are trained with encrypted images, and encrypted images are also used as test ones.
In this paper, one of the block-wise image encryption methods presented in \cite{aprilpyone2021block} is used to generate encrypted images.
The procedure of the encryption with pixel shuffling is summarized below (see Fig. \ref{fig:image_transformation})

\begin{figure}[ht]
    \begin{tabular}{cc}
        \begin{minipage}[b]{0.45\linewidth}
            \centering
            \includegraphics[keepaspectratio, scale=0.45]{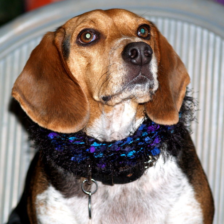}
        \end{minipage}&
        \begin{minipage}[b]{0.45\linewidth}
            \centering
            \includegraphics[keepaspectratio, scale=0.45]{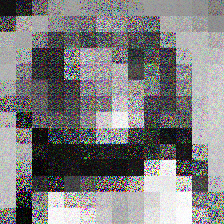}
        \end{minipage}\\
        \begin{minipage}[b]{0.45\linewidth}
            \centering
            \includegraphics[keepaspectratio, scale=0.45]{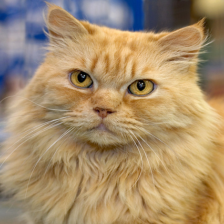}
        \end{minipage}&
        \begin{minipage}[b]{0.45\linewidth}
            \centering
            \includegraphics[keepaspectratio, scale=0.45]{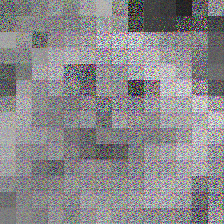}
        \end{minipage} \\
        Plain image & Encrypted image $(M=16)$
    \end{tabular}
    \caption{Example of encrypted images \label{fig:example_of_images}}
\end{figure}

\begin{enumerate}
    \item Split a three channel (RGB) color image $x$ into non-overlapped blocks with a size of $M \times M$ such that $ \{ B_{1 \, 1}, \dots , B_{w \, h}, \dots, B_{W_{\mathrm{b}} \, H_{\mathrm{b}}} \}$.
    \item Flatten each block $B_{w \, h}$ into a vector $b_{w \, h}$.
    \item Generate a secret key $K_i$ as
    \begin{equation}
        K_i = [ v_1, v_2, \dots, v_k, \dots, v_{k'}, \dots, v_{p_{\mathrm{b}}} ], 
    \end{equation}
    where $k, k' \in \{1, ..., p_{\mathrm{b}} \}$, and $v_k \neq v_{k'}$ if $k \neq k'$.
    \item Randomly permute pixels in each vector $b_{w \, h}$ to generate an encrypted vector $b'_{w \, h}$ as
    \begin{equation}
        b'_{w \, h}(k) = b_{w \, h}(v_k).
    \end{equation}
    \item Reshape each encrypted vector $b'_{w \, h}$ into an encrypted block $B'_{w \, h}$.
    \item Concatenate encrypted blocks into an encrypted image $x'$.
\end{enumerate}
Fig. \ref{fig:example_of_images} shows an example of encryption with the above procedure where $M=16$.
In the method, training and test images are encrypted by using the same key set.

\Figure[t][scale=0.17]{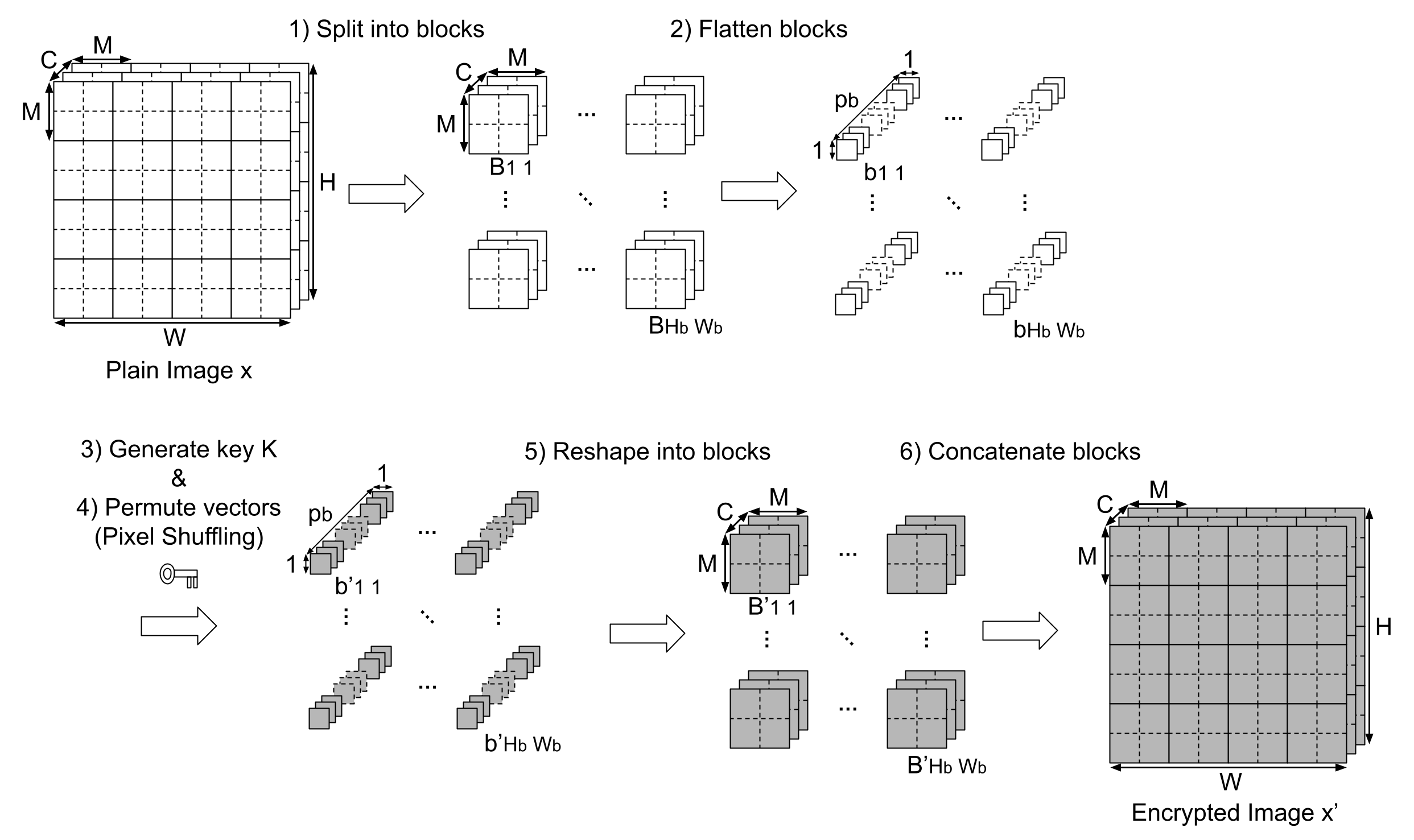}{Procedure of image encryption \label{fig:image_transformation}}

\par
\noindent
\textbf{Training Sub-models with Encrypted Images} \par
$N$ sub-models are fine-tuned by using images encrypted with $N$ keys as shown in Fig. \ref{fig:framework}.
The details of training sub-models are illustrated in Fig. \ref{fig:submodel}.
As shown in the figure, all training images are encrypted by using  key $K_i$, and then the pre-trained
ViT \cite{dosovitskiy2021an} is fine-tuned to produce an encrypted sub-model $i$ with the encrypted images.
$N$ sub-models are created with $N$ keys in accordance with the above procedure.

\Figure[t][scale=0.24]{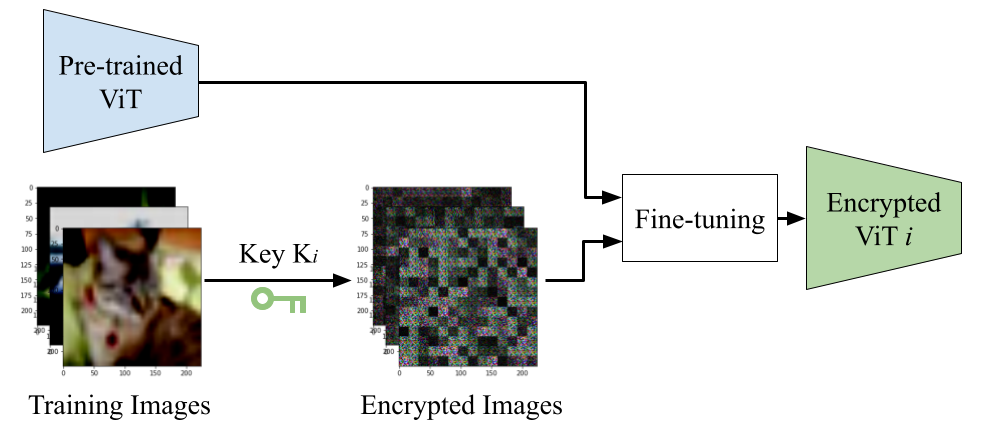}{Training of sub-models with encrypted images \label{fig:submodel}}

\par
\noindent
\textbf{Random Ensemble of Sub-models and Testing}
Fig. \ref{fig:architecture} shows the details of testing with a random ensemble of $N$ encrypted sub-models.
The steps for getting an estimation result from a test image are summarized below.
\begin{enumerate}
    \item Generate $N$ images encrypted with $N$ keys from a test image.
    \item Input each encrypted image into the corresponding sub-model.
    \item Select $S$ outputs ($3 \leq S \leq N$) from $N$ sub-models randomly.
    \item Determine a final output by the average of $S$ results.
\end{enumerate}
By using a random selection of $S$ outputs, the random ensemble model provides a different estimation value every time, even if the same image is input to the model.
Some black-box attacks such as Square Attack \cite{andriushchenko2020square} repeatedly input images to the model and gradually update the perturbations on the basis of the information obtained from prediction results.
Accordingly, it is expected that random ensemble models make such attacks more difficult.

\section{Experiment} \label{sec:experiment}
To verify the effectiveness of the proposed defense, we ran several experiments on the CIFAR-10 and ImageNet datasets in an image classification task.

\subsection{Experimental Setup}
Experiments were conducted on the CIFAR-10 \cite{krizhevsky2009learning} and ImageNet \cite{ILSVRC15} datasets.
CIFAR-10 comprises $60,000$ images with $10$ classes ($6,000$ images for each class), with $50,000$ images for fine-tuning and $10,000$ images for testing.
ImageNet consists of $1.28$M images for training and $50,000$ images for validation with $1000$ classes.
All images were resized to $224 \times 224 \times 3$ to fit the input to ViT and scaled to $[0, 1]$ as a range of values. \par

We used a pre-trained ViT with a patch size of $M = 16$, which was prepared in \cite{dosovitskiy2021an}, as sub-models where it was pre-trained with ImageNet-21k.
ImageNet-21k is a dataset consisting of $21,000$ classes with a total of $1,400$ million images, which were resized to an image size of $224 \times 224 \times 3$ when pre-training ViT.
For CIFAR-10, we fine-tuned ViT for $5,000$ epochs with the stochastic gradient descent (SGD) optimizer, which was implemented in PyTorch.
The parameters of the SGD optimizer were a learning rate of $0.03$ and a momentum of $0.9$.
For ImageNet, we also used the SGD optimizer and ran $20,000$ epochs for fine-tuning.
The parameters of the SGD optimizer were the same as for CIFAR-10.
For model encryption, pixel shuffling with a block size of $M = 16$, as the patch size of ViT, was used. \par

The robust models were evaluated by using AutoAttack (AA) \cite{croce2020reliable} which consists of three white-box attacks (APGD-ce, APGD-t, FAB-t) and one black-box attack (Square) as shown in Table \ref{tab:autoattack}.
The maximum perturbations for AA were ${\ell}_{\infty}, \epsilon=8/255$ on the CIFAR-10 dataset and ${\ell}_{\infty}, \epsilon=4/255$ on the ImageNet dataset.

\subsection{Image Classification Accuracy}
The effectiveness of random ensemble models was verified on CIFAR-10 in terms of image classification accuracy.
In Table \ref{tab:CIFAR_4}, four kinds of models were compared against five attacks, where ``clean'' means that test images did not include any adversarial noise, and ``simple ensemble'' means models without random selection \cite{maungmaung2021ensemble}.
From Table \ref{tab:CIFAR_4}, the experimental results are summarized as follows.
\begin{itemize}
    \item \textbf{ViT (baseline): }The baseline model that was trained with clean images achieved the highest clean accuracy.
    However, it was vulnerable to all attacks.
    \item \textbf{Encrypted ViT: }The model trained with encrypted images, which did not consist of sub-models, was robust against white-box attacks such as APGD-ce as confirmed in \cite{tanaka2022on}, but it was not robust against Square Attack and AutoAttack.
    \item \textbf{Simple Ensemble: }The ensemble model without random selection was still vulnerable to Square Attack and AutoAttack, but it had a higher clean accuracy than that of the encrypted ViT.
    \item \textbf{Random Ensemble: }The proposed model had a high classification accuracy even if white-box attacks were applied as well as for simple ensemble.
    In addition, it was robust against Square Attack and AutoAttack.
\end{itemize}
From Table \ref{tab:CIFAR_4}, it was confirmed that the use of encrypted ViT models was effective at improving robustness against white-box attacks.
In addition, by combining a random ensemble with encrypted sub-models, the model was robust against not only white-box attacks but also Square Attack.
As a result, it had a high accuracy even when using AutoAttack.
Table \ref{tab:ImageNet_1} shows experiment results on the ImageNet dataset.
It was confirmed that the trend in classification accuracy on ImageNet is similar to that on CIFAR-10.

\renewcommand{\arraystretch}{1.2}
\begin{table*}[h]
    \centering
    \caption{Classification accuracy (\%) on CIFAR-10 against AEs (${\ell}_{\infty}, \epsilon=8/255$).}
    \begin{tabular}{c|c|c|c|ccccc}
    \hline
    \multicolumn{3}{c|}{Source \& Target} & \multirow{2}{*}{Clean(\%)} & \multicolumn{5}{c}{Attack Method} \\
    \cline{1-3} \cline{5-9}
    Model & $N$ & $S$ & & APGD-ce (\%) & APGD-t (\%)& FAB-t (\%) & Square (\%) & AA (\%)\\
    \hline
    \hline
    ViT & \multirow{2}{*}{1} & \multirow{2}{*}{1} & \multirow{2}{*}{99.03} & \multirow{2}{*}{0.00} & \multirow{2}{*}{0.00} & \multirow{2}{*}{0.03} & \multirow{2}{*}{0.85} & \multirow{2}{*}{0.00} \\
    (baseline) & & & & & & & & \\
    \hline
    \multirow{2}{*}{Encrypted ViT} & \multirow{2}{*}{1} & \multirow{2}{*}{1} & \multirow{2}{*}{97.69} & \multirow{2}{*}{93.78} & \multirow{2}{*}{94.01} & \multirow{2}{*}{97.49} & \multirow{2}{*}{0.43} & \multirow{2}{*}{0.43} \\
    & & & & & & & &  \\
    \hline
    \multirow{2}{*}{Simple Ensemble}& \multirow{2}{*}{4} & \multirow{2}{*}{4} & \multirow{2}{*}{98.27} & \multirow{2}{*}{98.24} & \multirow{2}{*}{98.18} & \multirow{2}{*}{98.27} & \multirow{2}{*}{1.54} & \multirow{2}{*}{1.54} \\
    & & & & & & & &  \\
    \hline
    Random Ensemble& \multirow{2}{*}{4} & \multirow{2}{*}{3 or 4} & \multirow{2}{*}{98.20} & \multirow{2}{*}{98.22} & \multirow{2}{*}{98.23} & \multirow{2}{*}{98.33} & \multirow{2}{*}{\textbf{76.50}} & \multirow{2}{*}{\textbf{76.29}} \\
    (Ours) & & & & & &  & &  \\
    \hline
    \end{tabular}
    \label{tab:CIFAR_4}
\end{table*}
\renewcommand{\arraystretch}{1}

\renewcommand{\arraystretch}{1.2}
\begin{table*}[h]
    \centering
    \caption{Classification accuracy on ImageNet against AEs (${\ell}_{\infty}, \epsilon=4/255$).}
    \begin{tabular}{c|c|c|c|ccccc}
    \hline
    \multicolumn{3}{c|}{Source \& Target} & \multirow{2}{*}{Clean (\%)} & \multicolumn{5}{c}{Attack Method} \\
    \cline{1-3} \cline{5-9}
    Model & $N$ & $S$ & & APGD-ce (\%) & APGD-t (\%) & FAB-t (\%) & Square (\%) & AA (\%) \\
    \hline
    \hline
    \multirow{2}{*}{Simple Ensemble} & \multirow{2}{*}{4} & \multirow{2}{*}{4} & \multirow{2}{*}{79.82} & \multirow{2}{*}{79.83} & \multirow{2}{*}{79.83} & \multirow{2}{*}{79.82} & \multirow{2}{*}{7.37} & \multirow{2}{*}{7.37} \\
    & & & & & & & & \\
    \hline
    Random Ensemble & \multirow{2}{*}{4} & \multirow{2}{*}{3 or 4} & \multirow{2}{*}{79.75} & \multirow{2}{*}{79.76} & \multirow{2}{*}{79.77} & \multirow{2}{*}{79.78} & \multirow{2}{*}{\textbf{67.01}} & \multirow{2}{*}{\textbf{66.33}} \\
    (Ours) & & & & & & & & \\
    \hline
    \end{tabular}
    \label{tab:ImageNet_1}
\end{table*}
\renewcommand{\arraystretch}{1}

\subsection{Effects of Number of Sub-models}
In Tables \ref{tab:CIFAR_4} and \ref{tab:ImageNet_1}, four sub-models ($N=4$) were used to construct ensemble models.
In Table \ref{tab:CIFAR_5}, $N=5$ was used, and an experiment was carried out on the CIFAR-10 dataset.
From the table, the random ensemble model with $N=5$ was demonstrated to improve classification accuracy for Square Attack and AutoAttack, compared with the random ensemble model with $N=4$. \par
In contrast, the simple ensemble model with $N=5$ had almost the same accuracy as that of the simple ensemble model with $N=4$.
From these results, increasing the number of sub-models can improve the robustness against black-box attacks under only the use of a random ensemble of sub-models because a greater number of sub-models makes black-box attacks more difficult when using a random ensemble.

\renewcommand{\arraystretch}{1.2}
\begin{table*}[h]
    \centering
    \caption{Classification accuracy (\%) on CIFAR-10 against AEs (${\ell}_{\infty}, \epsilon=8/255, N=4 \, \mathrm{and} \, 5$).}
    \begin{tabular}{c|c|c|c|ccccc}
    \hline
    \multicolumn{3}{c|}{Source \& Target} & \multirow{2}{*}{Clean(\%)} & \multicolumn{5}{c}{Attack Method} \\
    \cline{1-3} \cline{5-9}
    Model & $N$ & $S$ & & APGD-ce (\%) & APGD-t (\%)& FAB-t (\%) & Square (\%) & AA (\%)\\
    \hline
    \hline
    \multirow{4}{*}{Simple Ensemble} & \multirow{2}{*}{4} & \multirow{2}{*}{4} & \multirow{2}{*}{98.27} & \multirow{2}{*}{98.24} & \multirow{2}{*}{98.18} & \multirow{2}{*}{98.27} & \multirow{2}{*}{1.54} & \multirow{2}{*}{1.54} \\
    & & & & & & & & \\
    \cline{2-9}
    & \multirow{2}{*}{5} & \multirow{2}{*}{5} & \multirow{2}{*}{98.29} & \multirow{2}{*}{98.23} & \multirow{2}{*}{98.24} & \multirow{2}{*}{98.29} & \multirow{2}{*}{1.68} & \multirow{2}{*}{1.68} \\
    & & & & & & & &  \\
    \hline
    & \multirow{2}{*}{4} & \multirow{2}{*}{3 or 4} & \multirow{2}{*}{98.20} & \multirow{2}{*}{98.22} & \multirow{2}{*}{98.23} & \multirow{2}{*}{98.33} & \multirow{2}{*}{\textbf{76.50}} & \multirow{2}{*}{\textbf{76.29}} \\
    Random Ensemble & & & & & & & &  \\
    \cline{2-9}
    (Ours) & \multirow{2}{*}{5} & \multirow{2}{*}{3, 4, or 5} & \multirow{2}{*}{98.23} & \multirow{2}{*}{98.18} & \multirow{2}{*}{98.21} & \multirow{2}{*}{98.29} & \multirow{2}{*}{\textbf{80.08}} & \multirow{2}{*}{\textbf{79.85}} \\
     & & & & & & & & \\
    \hline
    \end{tabular}
    \label{tab:CIFAR_5}
\end{table*}
\renewcommand{\arraystretch}{1}

\subsection{Comparison with State-of-the-Art}
Various adversarial defenses have been studied so far, and they have been compared to evaluate the performance of each defense method.
RobustBench is a standardized benchmark for adversarial robustness, and the goal of RobustBench is to systematically track the real progress in adversarial robustness \cite{croce2021robustbench}.
AutoAttack, which is an ensemble of white-box and black-box attacks, is used to standardize the evaluation.
Accordingly, we compared our method with state-of-the-art defenses under AutoAttack. \par

Table \ref{tab:CIFAR_2} shows the clean and robust accuracy of the top 5 models in RobustBench on the CIFAR-10  dataset, and the top 5 models on the ImageNet dataset are given in Table \ref{tab:ImageNet_2}.
They were evaluated by AA (${\ell}_{\infty}, \epsilon=8/255$) in Table \ref{tab:CIFAR_2} and AA (${\ell}_{\infty}, \epsilon=4/255$) in Table \ref{tab:ImageNet_2}, respectively.
From the tables, our models with a random ensemble achieved the highest accuracy in both clean and AutoAttack on the two datasets.

\renewcommand{\arraystretch}{1.2}
\begin{table}[h]
    \centering
    \caption{Comparison with other robust models in RobustBench \cite{croce2021robustbench} on CIFAR-10}
    \begin{tabular}{c|cc}
    Model & Clean & AA \\
    \hline
    \hline
    RaWideResNet-70-16 \cite{peng2023robust} & 93.27 & 71.07 \\
    \hline
    WideResNet-70-16 \cite{wang2023better} & 93.25 & 70.69 \\
    \hline
    ResNet-152 + WideResNet-70-16 & \multirow{2}{*}{95.23} & \multirow{2}{*}{68.06} \\
    + mixing network \cite{bai2023improving} &  & \\
    \hline
    WideResNet-28-10 \cite{cui2023decoupled} & 92.16 & 67.73 \\
    \hline
    WideResNet-28-10 \cite{wang2023better} & 92.44 & 67.31 \\
    \hline
    \textbf{Ours ($N=4, \, S=3 \, \mathrm{or} \, 4$)} & \textbf{98.20} & \textbf{76.29} \\
    \hline
    \textbf{Ours ($N=5, \, S=3, 4 \, \mathrm{or} \, 5$)} & \textbf{98.23} & \textbf{79.85} \\
    \hline
    \end{tabular}
    \label{tab:CIFAR_2}
\end{table}
\renewcommand{\arraystretch}{1}

\renewcommand{\arraystretch}{1.2}
\begin{table}[h]
    \centering
    \caption{Comparison with other robust models in RobustBench \cite{croce2021robustbench} on ImageNet}
    \begin{tabular}{c|cc}
    Model & Clean & AA \\
    \hline
    \hline
    Swin-L \cite{liu2023comprehensive} & 78.92 & 59.56 \\
    \hline
    ConvNeXt-L \cite{liu2023comprehensive} & 78.02 & 58.48 \\
    \hline
    ConvNeXt-L + ConvStem \cite{singh2023revisiting} & 77.00 & 57.70 \\
    \hline
    Swin-B \cite{liu2023comprehensive} & 76.16 & 56.16 \\
    \hline
    ConvNeXt-B + ConvStem \cite{singh2023revisiting} & 75.90 & 56.14 \\
    \hline
    \textbf{Ours ($N=4, \, S=3 \, \mathrm{or} \, 4$)} & \textbf{79.75} & \textbf{66.33} \\
    \hline
    \end{tabular}
    \label{tab:ImageNet_2}
\end{table}
\renewcommand{\arraystretch}{1}

\subsection{Discussion}
In the above experiments, we confirmed that our models with a random ensemble are robust against both white-box and black-box attacks.
In addition, if we can use a greater number of sub-models, the random ensemble models are more robust against black-box attacks.
We assume that the adversary knows the model architecture and has access to pre-trained models and training data.
In addition, we assume that the adversary also knows the mechanism of the key-based defense, but not the secret keys. \par
To clearly show the properties of our models, we investigated the relationship between the number of leaked keys and the robustness.
To examine the relationship, we conducted an additional experiment on the CIFAR-10 dataset (see Table \ref{tab:leakedkey}).
In the experiment, random ensemble models with $N = 4$ were evaluated under the use of APGD-ce.
In Table \ref{tab:leakedkey}, ``\# of leaked keys'' indicates the number of secret keys known to the adversary.
As shown in Table \ref{tab:leakedkey}, random ensemble models achieved high accuracy when the number of leaked keys was zero or one.
In contrast, for encrypted models without any sub-model \cite{maung2020encryption,aprilpyone2021block}, which have one key, the accuracy will be significantly
reduced.
Accordingly, the use of sub-models can also enhance robustness against the leak of keys.

\renewcommand{\arraystretch}{1.2}
\begin{table}[t]
    \centering
    \caption{Robustness of random ensemble model against key leaks}
    \begin{tabular}{c|c|c|c|c}
    \hline
    \multicolumn{3}{c|}{Source \& Target} & \multirow{2}{*}{\# of leaked keys} & Attack Method \\
    \cline{1-3} \cline{5-5}
    Model & $N$ & $S$ & & APGD-ce \\
    \hline
    & \multirow{4}{*}{4} & \multirow{4}{*}{3 or 4} & 4 & 0.26 \\
    Random & & & 2 & 27.86  \\
    Ensemble & & & 1 & 95.65  \\
    & & & 0 & 98.22 \\
    \hline
    \end{tabular}
    \label{tab:leakedkey}
\end{table}
\renewcommand{\arraystretch}{1}

\section{Conclusion} \label{sec:conlusion}
In this paper, we proposed a novel method for adversarial defenses.
In the method, a random ensemble of ViTs encrypted with secret keys is used to construct a model.
Simple encrypted models are known to be robust against white-box attacks, but not against black-box ones.
To enhance robustness against all attacks, the use of ViT and an ensemble of sub-models was proposed in the paper.
In addition, a benchmark attack method, called AutoAttack, which consists of four attacks: Auto-PGD-cross entropy, APGD-target, FAB-target, and Square Attack, was applied to models to test adversarial robustness objectively.
In an experiment, the proposed method was demonstrated to be robust against not only white-box attacks but also black-box ones in an image classification task.
In addition, it was compared with state-of-the-art defenses in a standardized benchmark for adversarial robustness, RobustBench, and it was verified to outperform the conventional ones in terms of clean accuracy and robust accuracy.

\bibliographystyle{IEEEtran}
\bibliography{main}

\EOD

\end{document}